\pdfoutput=1

\documentclass[11pt]{article}

\usepackage{acl}

\usepackage{times}
\usepackage{latexsym}

\usepackage[T1]{fontenc}

\usepackage[utf8]{inputenc}

\usepackage{microtype}

\usepackage{inconsolata}

\usepackage{CJKutf8}
\usepackage{tcolorbox}
\usepackage{xcolor}
\usepackage{hyperref}
\usepackage{graphicx}
\usepackage{xspace}
\usepackage{multirow}
\usepackage{url}
\usepackage{algorithm}
\usepackage{algorithmicx}
\usepackage{algpseudocode}

\newcommand{\uu}{\( u \)\xspace}

\title{
    \begin{tabular}{@{}l@{\hspace{10pt}}c@{}}  
        \raisebox{-0.25\height}{\includegraphics[width=1.3cm]{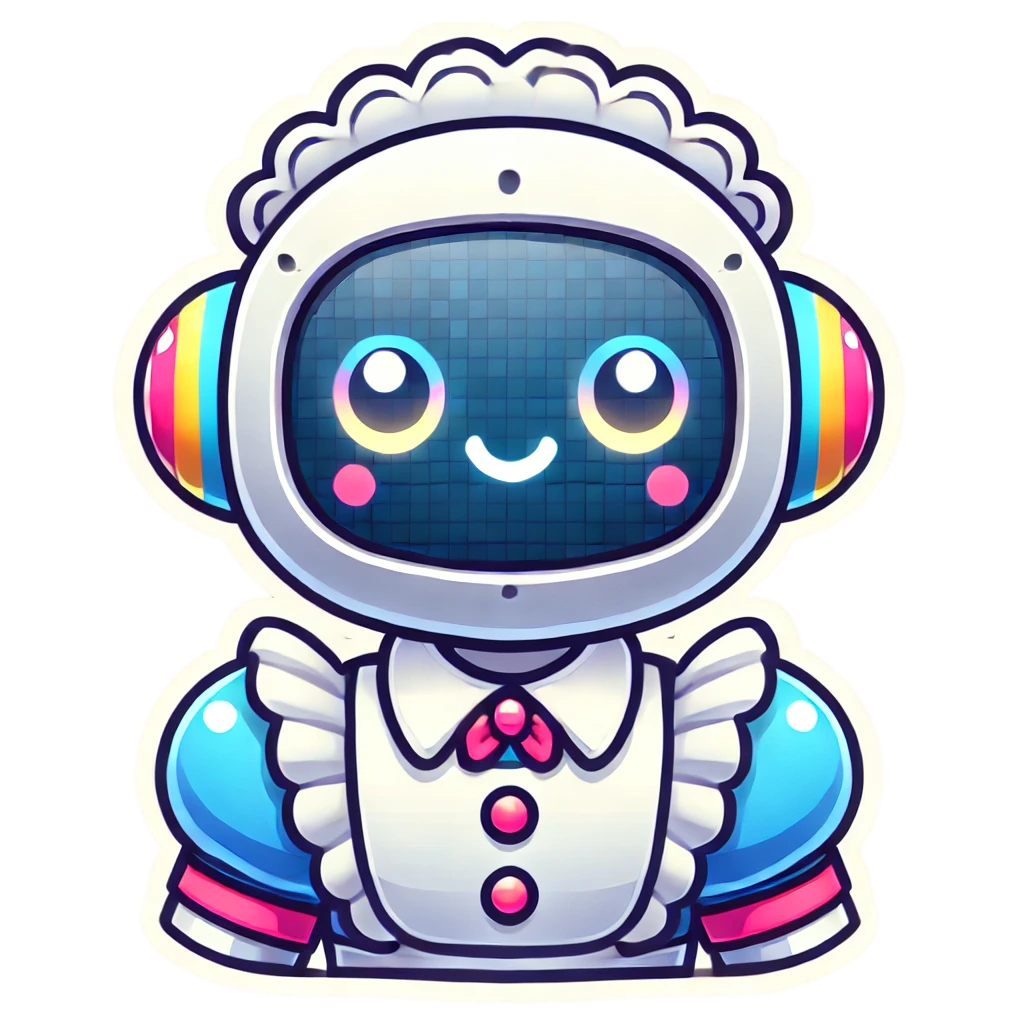}} & 
        {\textsc{AI Persona}:Towards Life-long Personalization of LLMs}
    \end{tabular}
}

\author{
 \textbf{Tiannan Wang\textsuperscript{1}\thanks{equal contribution.}},
 \textbf{Meiling Tao\textsuperscript{2$*$}},
 \textbf{Ruoyu Fang\textsuperscript{3}},
 \textbf{Huilin Wang\textsuperscript{4}},
 \textbf{Shuai Wang\textsuperscript{5}}\\
 \textbf{Yuchen Eleanor Jiang\textsuperscript{1}},
 \textbf{Wangchunshu Zhou\textsuperscript{1}},
\\
\\
 \textsuperscript{1}OPPO AI Center, 
 \textsuperscript{2}Guangdong University of Technology \\
 \textsuperscript{3}University of Illinois at Urbana-Champaign, 
 \textsuperscript{4}Beihang University, 
 \textsuperscript{5}Tsinghua University
\\
}

\begin{document}
\maketitle
\begin{abstract}
In this work, we introduce the task of life-long personalization of large language models. While recent mainstream efforts in the LLM community mainly focus on scaling data and compute for improved capabilities of LLMs, we argue that it is also very important to enable LLM systems, or language agents, to continuously adapt to the diverse and ever-changing profiles of every distinct user and provide up-to-date personalized assistance. We provide a clear task formulation and introduce a simple, general, effective, and scalable framework for life-long personalization of LLM systems and language agents. To facilitate future research on LLM personalization, we also introduce methods to synthesize realistic benchmarks and robust evaluation metrics. We will release all codes and data for building and benchmarking life-long personalized LLM systems. \footnote{\url{https://github.com/tml1026/Lifelong-Personalized-Agent}}
\end{abstract}

\section{Introduction}
The advent of large language models (LLMs) that can effectively understand and generate human language~\citep{gpt,gpt2,gpt3,instructgpt,gpt4,touvron2023llama,llama2} has marked a transformative era in artificial intelligence. LLMs such as ChatGPT, Gemini, and Claude have shown remarkable capabilities, not only as general chatbot providing useful responses but also as agents~\citep{zhou2023agents,zhou2024agents2,chen2023agentverse,liang2024self,liang2024cmat} that assist hundreds of millions of users in diverse real-world tasks. Mainstream efforts in the AI/LLM community have been focusing on scaling data and compute on pre-training or post-training stages for building stronger LLMs that can better serve the needs of users. Millions of GPUs and billions of dollars are consumed every year for this purpose.

However, a crucial question is: \textit{does satisfactory user experience naturally emerges with improved capabilities of LLMs?} While it is true that improved reasoning abilities help LLMs solve more complex tasks and improved instruction-following abilities enable LLMs to better understand user intents, current LLM systems are by design incapable of capturing diverse and ever-changing personal profiles of distinct users that are implicitly encoded in life-long human-AI interaction histories. For example, an AI assistant must have rich information of a user's residential address, personal agenda, income and consumption habits, preferences for foods and restaurants, etc., to generate a satisfactory response for a query as simple as "help me reserve a restaurant for dinner."  Moreover, many aspects, including personalities, intents, preferences, etc., in user profiles are \textit{dynamic} and \textit{ever-changing} in the real world, making it crucial for LLM systems to be capable of constantly adjusting to the changes of user profiles.
Therefore, we argue that stronger capabilities of LLMs are \textit{not} all we need for AGI and \textit{life-long personalization} of LLMs is another important building block for AI systems that are helpful, satisfactory, and engaging for everyone.

While personalization has been carefully investigated in other domains such as recommendation systems~\citep{barkan2020explainable,woźniak2024personalized,dai2024language,li2024personal,yang2023palr,kang2023llms}, research on LLM personalization is quite limited and suffers from three major limitations: First, most recent work on LLM personalization~\citep{mysore2023pearl,zhiyuli2023bookgpt,yang2023refgpt} are task-specific and the methodology for LLM personalization can not be generalized to other tasks; Second, most existing LLM personalization methods requires training either the entire LLM or a few blocks within the LLM, making them impossible or very costly to scale to real-world applications used by million of users on a daily basis; Finally, the lack of diverse and realistic benchmarks makes it hard to evaluate new methods. To be specific, most recent work conducts experiments on the LaMP benchmark~\citep{salemi2024lamp}. However, the tasks in this benchmark such as citation identification and scholarly title generation are very different from queries asked by real-world users and therefore not representative enough. Moreover, recent study~\citep{salemi2024optimization} has shown that a non-personalized LLMs can also achieve competitive performance on these tasks, suggesting that the benchmark is not suitable for evaluating advanced LLM personalization approaches. In addition, all previous studies consider LLM personalization as a one-time task. However, it is crucial for LLM systems to constantly adjusting to ever-changing user profiles in real-world applications by learning from human-AI interactions.

In this paper, we introduce the task of life-long personalization of LLMs and provide a detailed task formulation emphasizing the difference from previous \textit{static} LLM personalization problems. We present \textsc{AI Persona}, a simple, general, effective, and scalable framework to build life-long personalized LLM systems or language agents. Specifically, we define user profiles as learnable dictionaries where the keys are fields representing various aspects of a real-world user, including demographics, personality, usage patterns, and preferences. The values are the users' personal information in the corresponding fields. During inference, a user's AI persona will be dynamically assembled into a part of the prompt/input to the LLM backbone so that it will generate personalized responses. Life-long adaptation of user profiles is achieved by a carefully designed LLM-based persona optimizer, which constantly adjust the AI Persona of the user during the interaction between the user and the LLM system. To the best of our knowledge, the proposed framework is the first method in the literature that can constantly adjust the profiles of each users during the progress of human-AI interaction. Moreover, our approach does not involve model training and only requires to store a lightweight config file for each user, making it scalable for real-world applications with large amount of users.

To test the effectiveness of the proposed \textsc{AI Persona} framework and facilitate future research, we build \textsc{PersonaBench}, a benchmark for (life-long) LLM personalization research. \textsc{PersonaBench} consists of diverse and realistic user profiles and user queries generated with a carefully designed LLM-based workflow. Our experiments on \textsc{PersonaBench} demonstrates that the proposed \textsc{AI Persona} framework\footnote{We use AI Persona and Persona Learning interchangeably in this paper.} can effectively learn and adapt to user profiles over time.

Our contributions can be summarized as follows:

\begin{enumerate}
    \item We provide a clear definition of life-long personalization, emphasizing the necessity of continuous adaptation in understanding user needs. To our knowledge, this is the first work to address this task in the literature.
    
    \item We propose a novel pipeline for generating realistic persona chat data, encompassing diverse persona configurations and authentic user-agent interaction simulations.
    
    \item We introduce a life-long personalized agent framework, serving as a baseline solution for this task.
    
    \item We conduct experiments that demonstrate the effectiveness of our methods, showcasing significant improvements in agent performance and adaptability to user personas.
\end{enumerate}

By addressing these aspects, our work aims to advance the field of personalized conversational agents, fostering more effective and meaningful user interactions.

\section{Related Work}

\subsection{Personalized LLMs}
The personalization of language models (LLMs) has garnered significant interest in industries such as recommendation systems and search, focusing on providing tailored responses that adapt to individual user preference~\citep{barkan2020explainable,woźniak2024personalized,dai2024language,li2024personal,yang2023palr,kang2023llms,tan2023user,tao2023rolecraft}. Recently, this focus has extended to other domains such as research assistants~\citep{wang2024surveyagent,lin2024paper}, travel planners~\citep{xie2024travelplanner}, writing assistants~\citep{mysore2023pearl}, book recommending~\citep{zhiyuli2023bookgpt}, shopping counselors~\citep{yang2023refgpt}, and programming agents~\citep{gao2023assistgui}.

A common approach involves \textbf{fine-tuning personalized LLMs}. ~\citet{zhou-etal-2023-learning} integrate persona prediction with response generation, while~\citet{tan2024demo} use LoRA~\citep{hu2021lora} to fine-tune Llama~\citep{touvron2023llama} for individual users. To further enhance efficiency, ~\citet{zhuang2024hydra} and ~\citet{tan2024personalized} group users and fine-tune LoRA at the group level.

Despite these advancements, fine-tuning approaches face limitations in maintaining adaptability over time due to the need for frequent retraining, which is impractical in real-world scenarios. To address this, \textbf{RAG-based personalized LLMs}, offer an alternative by leveraging user-specific historical data. For example, ~\citet{salemi2024lamp} introduced a pseudo-RAG approach for incorporating user history, later enhanced by~\citet{salemi2024optimization} through retriever optimization.  Similarly,~\citet{li2023teach} and~\citet{mysore2023pearl} propose retrieval-based methods to integrate user-authored documents for prompt augmentation. However, the input length constraints still hinder effective personalization when user interactions are lengthy. To address this, some studies have utilized comprehensive user histories to generate summaries based on user interactions~\citep{christakopoulou2023large, zhiyuli2023bookgpt, richardson2023integrating}. 
While significant progress has been made, existing approaches rarely consider realistic, life-long adaptation scenarios, leaving room for further exploration in dynamic, long-term personalization.

\subsection{Benchmarking Personalized LLMs}
While benchmarks for LLM agents have been extensively developed~\citep{shen2023taskbench,mialon2023gaia,Liu2023AgentBench}, few focus on personalized agents. A significant challenge in this area is the scarcity of realistic data. Existing benchmarks, such as LaMP~\citep{salemi2024lamp}, utilize public datasets with user identifiers, offering limited user-data associations. LaMP comprises seven tasks, primarily binary classification and single-instance generation, such as \textit{Personalized Citation Identification} (choosing which paper is likely to be cited) and \textit{Personalized Scholarly Title Generation} (generating a title from an abstract). However, its minimal reliance on historical user data limits its ability to evaluate true personalization. Notably, these task designs are insufficient since a large language model (LLM) could achieve competitive performance without access to any historical user data about \( u \)~\cite{salemi2024optimization}.

To overcome these limitations, our proposed data synthesis pipeline generates a broad spectrum of persona configurations and user-agent interactions, simulating realistic conversational scenarios that evolve. This approach not only supports the development of more adaptable agents but also establishes a comprehensive benchmark for evaluating life-long personalization capabilities. By dynamically generating data that reflects diverse user behaviors and preferences, we provide a more robust framework for training and testing personalized LLMs in environments that closely resemble real-world usage.

\section{AI Persona: Towards Life-long Personalized LLMs}

\subsection{Task Formulation}
We will provide a clear definition and task formulation of \textbf{Life-long personalized LLM} in this section. 
Consider a model or an agent, noted \(P\), which takes the input query \(x\) from a user~\uu. A personalized agent should conditioning on not only \(x\) but also the profile of \uu in order to generate satisfactory responses that is tailored to this specific \uu. In previous setting, particularly in LaMP \citep{salemi2024lamp}, the profile of \uu, \(P_u\), is defined as a set of user’s historical data, i.e., the past input and personalized outputs produced by or approved by the user.
That is 
\( P_u = \{(x_{u1}, y_{u1}), (x_{u2}, y_{u2}), \ldots, (x_{uk}, y_{uk})\}\). Through out the evaluation, the \(P_u\) is fixed for each \uu which makes the persona profile static.
In this work, we re-define the user profile as learnable dictionaries where the keys are fields representing various aspects of a real-world user and the values are the users’ personal information or traits in the corresponding field. 

\[
P_u = \{(k_1, v_{u1}), (k_2, v_{u2}), \ldots, (k_n, v_{un})\},
\]

where \( k_i \) represents the \(i\)-th field of user attributes such as \textit{demographics}, \textit{personality}, \textit{usage patterns}, and \textit{preferences}, and \( v_{ui} \) denotes the corresponding value reflecting user \( u \)'s information in that field.

The learnable aspect of \(P_u\) means that each value \( v_{ui} \) is dynamically updated based on the ongoing interactions between the user and the agent \( P \). Mathematically, this is modeled as:
\[
v_{ui}^{(t)} = f_{\theta}(v_{ui}^{(t-1)}, (x_t, y_t)),
\]

where \( f_{\theta} \) is the persona optimizer parameterized by \(\theta\), and \((x_t, y_t)\) is the interaction data at time step \( t \). Note that \( f_{\theta} \) can either be a learnable model with trainable parameters or a pre-trained LLM-based agent guided by well-designed prompts. In this work, we opt for the latter approach, keeping \( \theta \) fixed and utilizing the LLM's emergent abilities to adapt through prompting rather than parameter updates.
This formulation allows \(P_u\) to continuously evolve, capturing the latest user behaviors and preferences in real-time. 

In contrast to static representations, our approach ensures that the user profile remains \textbf{up-to-date} and \textbf{context-aware}, enabling the model to adapt its responses more accurately to the evolving characteristics of the user in a longer life-span.

\subsection{Data Generation Pipeline}
\begin{figure}
    \centering
    \includegraphics[width=0.98\linewidth]{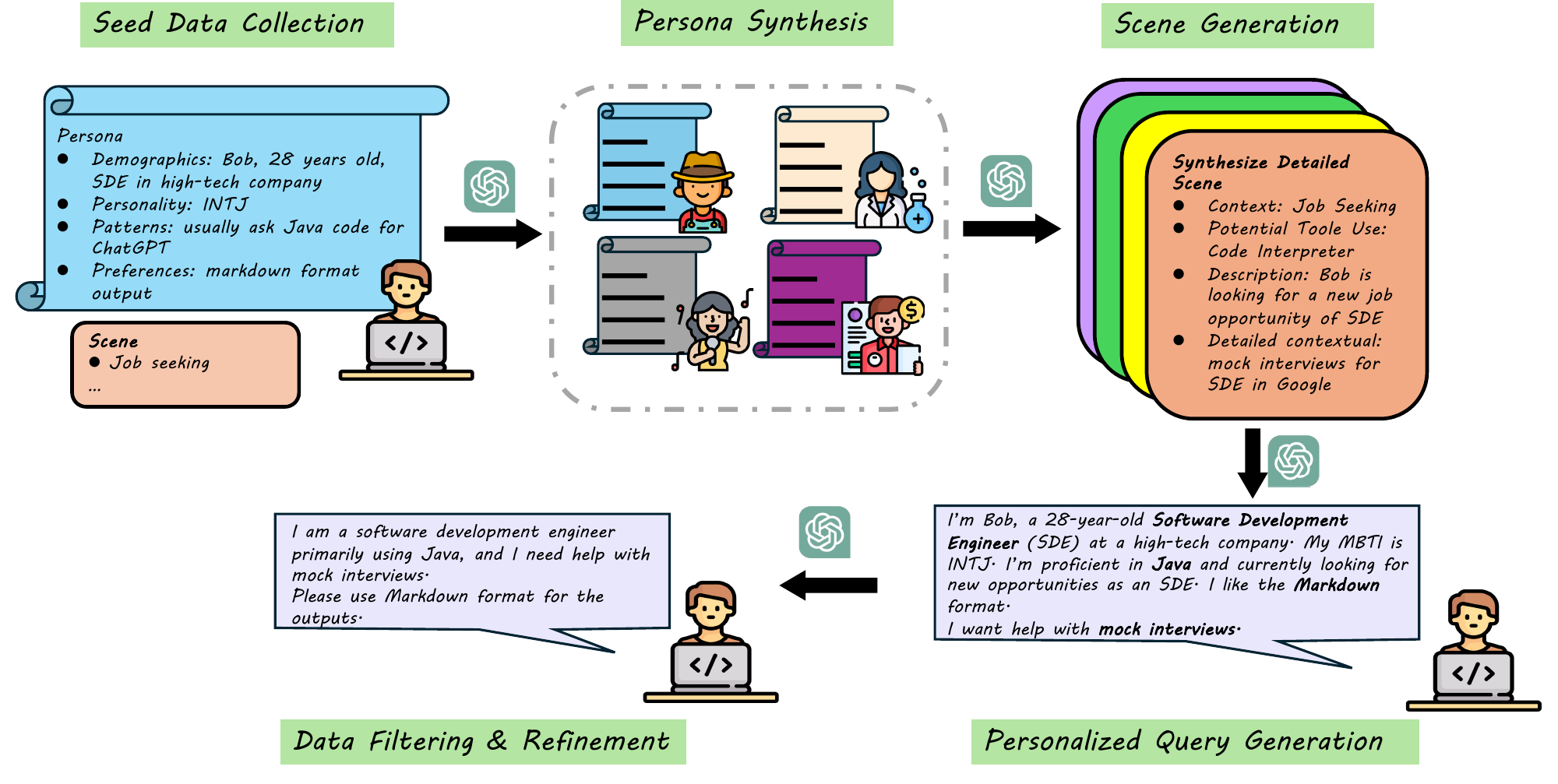}
    \caption{Data generation pipeline for PersonaBench. This pipeline consists of 5 stages: seed data collection, persona synthesis, scene generation, personalized query generation and data filtering and refinement.}
    \label{fig:data-generation}
\end{figure}
The most challenging aspect of personalization is the scarcity of realistic user data and the difficulty of accessing it. To tackle this problem, we propose a persona chat data generation pipeline, as shown in Figure\ref{fig:data-generation}, to synthesize real persona profile and generate user-agent conversation data according to each aspect of the persona profile.

\subsubsection{Persona Generation}
First we pre-define the necessary fields that a comprehensive persona profile should contain, which are \textit{demographics}, \textit{personality}, \textit{usage patterns}, and \textit{preferences}.
\begin{itemize}

\item \textbf{Demographics}: This field captures key factual information about the user's identity and background, including age, gender, nationality, language, and career information or education background. It provides a basic understanding of the user.
\item \textbf{Personality}: This field defines the user's psychological characteristics and values, reflecting how they typically think, feel, and behave. In our setting, we represent it using MBTI (Myers-Briggs Type Indicator) and interests. Personality traits implicitly influence how users express themselves, respond to different conversational styles, and engage with the agent.

\item \textbf{Patterns}: This field represents the user’s habits and interactions with the personalized agents, such as behavior engagement patterns, usage patterns and purchase pattern. Understanding usage patterns enables the agent to anticipate user needs and provide proactive support, thereby enhancing user experience and engagement.

\item \textbf{Preferences}: This field encompass the user's preferred interaction styles, formats, and workflows. Capturing preferences helps the agent to personalize responses and recommendations, making interactions more relevant and satisfying for the user.
\end{itemize}
To create more comprehensive and realistic persona profiles, we first ask volunteers from diverse backgrounds (e.g., different professions and life stages) who has a habit to regularly use AI products to complete persona profiles. Using these real personas as seed configurations, we then prompt the LLM to summarize these profiles into brief descriptions as seed hints. With the seed hints and seed configurations in place, we instruct the LLM to generate diverse persona descriptions in a self-instruct manner \citep{self-instruct} to generate a large amount of persona hints, then using the seed configurations as in-context exemplars to guide the model in producing diverse, comprehensive and realistic persona data. \footnote{We incorporate seed hints into the persona data generation process because we found that when instructing the LLM to generate configurations solely based on seed configurations, it tends to produce similar personas even when different exemplars and temperature are used during inference.}

\subsubsection{Scene generation}
In order to synthesize realistic user-agent chat data, a critical component is to generate realistic and personalized queries. Previous work only focus on chit-chat or role-play scenario \citep{jandaghi2023faithful, wang2023rolellm}. In contrast, our work aims at more practical settings where personalized agent is involved in solving real-world problems.

To achieve realistic query generation, we believe it is important to provide the LLM comprehensive and contextual information besides the persona information. Therefore,we first identify and define several common scenes in which people might utilize AI as an assistant. By prompting LLM to generate more detailed scene descriptions according to different persona profiles, these common scenes are then adapted to various personas, resulting in personalized scene descriptions that better reflect the unique preferences and characteristics of different users. Next, we use these common scenes as in-context exemplars, prompting the LLM to generate persona-specific scene descriptions based on the given persona profile.

To achieve realistic query generation, we first identify and define several common scenes in which people might utilize AI as an assistant. These common scenes serve as a baseline and are then adapted to various personas, resulting in personalized scene descriptions that better reflect the unique preferences and characteristics of different users. Next, we use these common scenes as in-context exemplars, prompting the LLM to generate persona-specific scene descriptions based on the given persona profiles.
\paragraph{Contextual Scene Information}
We prompt the LLM to enrich the scene descriptions with additional contextual information that is tailored to both the persona and the specific scenario. For instance, in the context of a \textit{Job Seeking} scene, the synthesized description for \textit{Brandon}, a virtual persona who is a master's student specializing in computer vision and deep learning, might involve preparing for campus recruitment. The contextual complement for this scene would include specific topics relevant to Brandon's field of study and career, such as \textit{mock interviews for CV engineer} or \textit{hot questions in high-tech companies}, etc. These detailed contextual elements help create a more realistic and personalized interaction, ensuring that the generated queries are both contextually relevant and aligned with the persona's background and goals.

\paragraph{Function Call Generation}
One of the major limitations of current LLMs compared to fictional AI assistants like Jarvis in the Iron Man movies is their inability to interoperate in the real world autonomously. To address this gap and make our setting more practical, we incorporate function call data into our benchmark. This is particularly important for scenarios where users expect the AI to execute specific tasks on their behalf, such as checking whether they should carry an umbrella today, planning the remaining budget for the month, or searching for job opportunities that match their profile. 
For each scene, we identify potential API functions that align with the user's intended actions, providing the personalized LLM with a structured way to obtain relevant information or perform the desired tasks.

\subsubsection{Personalized Query Generation}
After establishing personalized scene descriptions and their contextual information, the next step is to generate realistic and contextually appropriate queries. To achieve this, we employ a user simulator capable of role-playing based on a given persona profile. 
The user simulator reads the persona profile, current scene description, along with its contextual information to produce relevant and nuanced queries. This approach moves beyond generic role-play or simple chit-chat, instead generating queries that are deeply rooted in realistic scenarios that users may encounter.

\subsubsection{Data Filtering and Refinement}
To avoid generating unanswerable or nonsensical queries, the model evaluates whether it can provide a reasonable response, filtering out any queries that fail this criterion.
Additionally, to prevent the user simulator from directly revealing persona traits in the query, we conduct a data refinement procedure that neutralizes the query, retaining only the essential information and the intended purpose.

\subsection{AI Persona Framework}
Our proposed AI Persona framework is composed of three main components: a \textbf{Historical Session Manager}, a \textbf{Tool Executor}, and a \textbf{Personalized Chatbot}. Each component plays a critical role in enabling life-long personalized interactions with users.

\paragraph{Historical Session Manager}
The Historical Session Manager is responsible for managing and storing conversation histories across multiple sessions for different users. It provides a comprehensive record of user interactions, enabling the system to maintain context and continuity over time. Its core functionalities include initializing, loading, saving, and retrieving conversation sessions, ensuring that the system can seamlessly recall past interactions to support coherent and context-aware responses.

\paragraph{Tool Executor}
The Tool Executor is a well-prompted LLM designed to simulate external API execution. It interprets function calls from the Personalized Chatbot and generates appropriate responses based on predefined API descriptions provided in the scene information. Its primary function is to generate realistic and contextually accurate function call responses, bridging the gap between the chatbot and external tools or databases.

\paragraph{Personalized Chatbot}
The Personalized Chatbot is a well-designed conversational agent that leverages user personas to deliver personalized and context-aware responses. It acts as a versatile agent, capable of dynamically adapting its behavior to align with the user's evolving profile and current context. Its core functionalities include:

\begin{itemize} 
\item \texttt{Persona-based Interaction:} The chatbot generates tailored responses based on the user’s persona and query, ensuring that each interaction is relevant and engaging. 
\item \texttt{Dynamic Persona Updates:} The chatbot updates the user’s persona profile in real-time as the interaction progresses, reflecting any changes in user preferences, behaviors, or context. 
\item \texttt{Function Calling:} When necessary, the chatbot initiates appropriate function call to external APIs associated with the current scene, enriching the response with precise and contextually appropriate information. 
\end{itemize}

\begin{figure}
    \centering
    \includegraphics[width=0.9\linewidth]{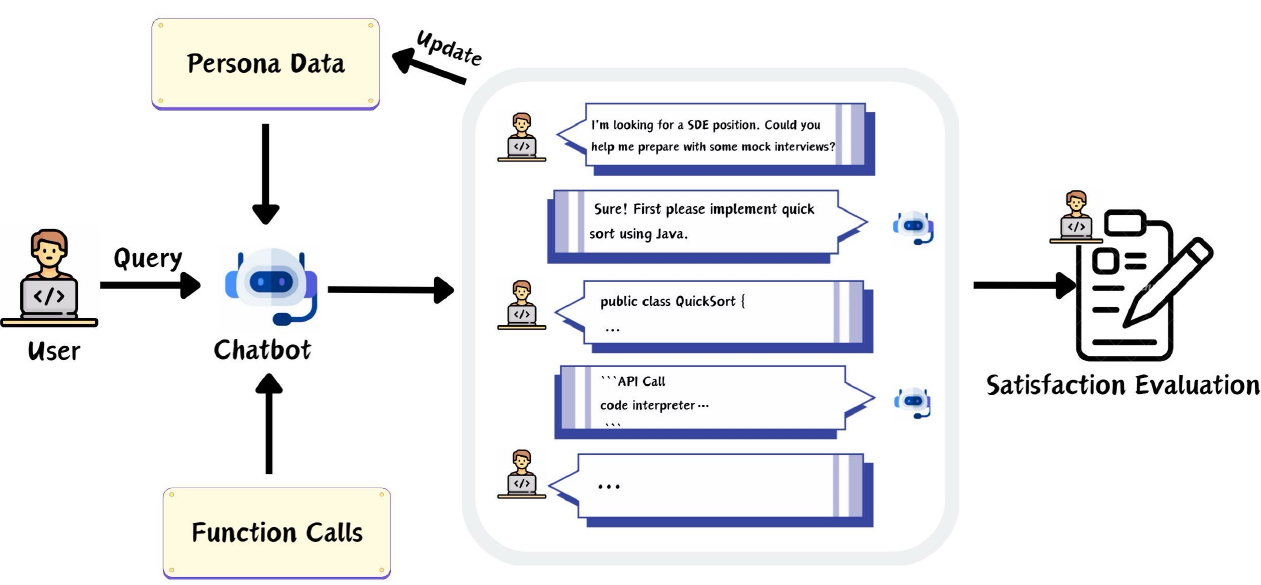}
    \caption{AI Persona Framework.}
    \label{fig:enter-label}
\end{figure}
As shown in Figure \ref{fig:enter-label}, during inference, the framework operates in a sequential manner, integrating all three components to provide a coherent and personalized user experience. Whole process is illustrated in the Algo~\ref{algo:inference process}.

\paragraph{Step 1: Persona and Session Initialization}
When a new conversation begins, the Personalized Chatbot loads the user’s persona profile(if exists) to model the persona and context. Concurrently, the user simulator—acting as the user—interprets the current scene and formulates a query based on the persona attributes and the scene description. This step initiates the session's conversation. 

\paragraph{Step 2: Query and Response Generation}
Based on the integrated context and scene understanding, the Personalized Chatbot then generates a tailored response by considering the user’s persona and the query. It may also issue function calls to the Tool Executor if external data or actions are required to enhance the response and, if necessary, explicitly includes these function calls in the response. For example, in a \textit{job-seeking} scenario, the chatbot might call an web\_search API to look for the latest interview questions for a specific role, providing user with informative advices.

\paragraph{Step 3: Tool Execution and Information Integration}
When the chatbot issues a function call, the Tool Executor interprets the request and simulates the external API execution according to pre-defined API documentations. It then returns the generated response, which the chatbot incorporates into its final output to the user. This allows the chatbot to provide an informative and accurate response, seamlessly integrating external data into the conversation.

\paragraph{Step 4: Satisfaction Evaluation}
After receiving the response, the User Simulator conducts a satisfaction evaluation to determine whether the generated response meets the expectation. This evaluation is based on a pre-generated reference response, an abstract expectation that represents the ideal outcome for the given persona and scenario. The User Simulator reviews the current session's conversation history and, by referencing both the persona configuration and the expected response, assesses whether the chatbot’s reply aligns with the user’s needs and objectives, as defined by the persona. 

\paragraph{Step 5: Persona Update and Session Storage}
When user satisfaction is confirmed, this conversation session is deemed finished. The personalized chatbot will then updates the user’s persona profile if necessary. For example, if the user expresses new preferences or changes their goals, these updates are reflected in the persona profile. Practically, the model updates the persona configuration after every \(k\) sessions, allowing it to accumulate more interaction data before making adjustments.
Finally, the Historical Session Manager saves the current session data, ensuring that all interactions are recorded for future reference.

This multi-step process enables the AI Persona Framework to provide nuanced and adaptive interactions that cater to the user's individual needs, fostering a more engaging and personalized user experience.

\begin{algorithm}[ht]
\caption{AI Persona Framework Inference Process}
\begin{algorithmic}
\State \textbf{Input:} Persona profile $P$, Scene information $S$, Personalized Chatbot $LM$ and User simulator $U$
\Statex
\textcolor{olive}{\# Initialization}
\State $H \gets$ Conversation history $[\textit{empty list}]$ 
\State $T \gets$ Tool Executor loads $S$
\State \texttt{satisfied} $\gets$ \textbf{False} 
\While {not \texttt{satisfied}} \\
    \textcolor{olive}{\# Query Generation}
    \State $Q \gets U$.get\_query($P$, $S$,  $H$)  \\
    \textcolor{olive}{\# Response Generation with tool executions}
    \State $R \gets LM$.get\_response($Q$, $P$, $H$, $T$)\\
    \textcolor{olive}{\# Append query-response pair to chat history}
    \State $H \gets H \cup \{(Q, R)\}$ \\
    \textcolor{olive}{\# Satisfaction Check}
    \State \texttt{satisfied} $\gets U$.satisfaction\_check($P$, $R$, $H$)
\EndWhile \\
\textcolor{olive}{\# Persona Update and Chat History Storage}
\State Update $P$ and save $H$ 

\end{algorithmic}
\label{algo:inference process}
\end{algorithm}

\section{Experiments}
\subsection{Experiment Setup}
\paragraph{Benchmark Setting}
Our proposed benchmark, \textsc{PersonaBench}, is composed of 200 diverse persona profiles, each paired with 10 common scene settings and 10 persona-specific scene settings. For each persona, we randomly sample 3 to 5 different scenes and prompt the LLM to re-generate scene descriptions, contextual information, and potential function calls. This approach simulates a common scenario where users often ask similar questions about the same topics over time. By incorporating this data type, we aim to assess whether a personalized LLM can learn from previous sessions and improve performance when handling similar scenarios. Additionally, we pre-generate the initial user queries and carefully hand-check each one to ensure a fair comparison. In total, we synthesized over 6,000 data points to construct the benchmark.

\paragraph{Baselines}
We compare our proposed AI Persona framework with 3 baselines:

\begin{itemize}
    \item \textbf{No Persona Access}: The model generates responses without any access to the persona configuration, simulating a scenario where the model serves as an AI chatbot.
    
    \item \textbf{Golden Persona Access}: The model is provided with access to the ground truth persona configuration during inference, enabling it to generate responses that are fully aligned with the user's defined attributes.
    
    \item \textbf{Conversations RAG}: The model could not maintain or learn the user's persona but it could retrieve conversations in which there exist similar queries to generate responses according to historical interactions. 
\end{itemize}

\paragraph{Evaluation Metrics}
We evaluate the models across several key dimensions to measure their ability to align with user personas and efficiently meet user needs:

\begin{itemize}
    \item \textbf{Persona Satisfaction}: We use an LLM as a judge to score the first utterance of each session based on how well the generated responses solve the problem and how well they align with the user's persona. This score reflects if a chatbot can instantly get users' intentions.
    
    \item \textbf{Persona Profile Similarity}: After the session ends, we evaluate the final saved persona profile by comparing it to the ground truth persona. This measure reflects how accurately the model has updated and maintained the persona throughout the interactions.
    
    \item \textbf{Utterance Efficiency}: We measure how many utterances are required for the model to fully satisfy the user's needs. Fewer utterances indicate better alignment and understanding of the user’s requirements, as the model can meet the user's needs more efficiently with less back-and-forth interaction.
\end{itemize}

\begin{table*}[ht]
\centering
\caption{Performance of different persona settings.}
\resizebox{\textwidth}{!}{%
\begin{tabular}{c|cccc}
\hline
\multirow{2}{*}{\textbf{Setting}} & \multicolumn{2}{c}{\textbf{Personalized Response}} & \textbf{Persona Similarity} & \textbf{Utterance Efficiency}\\ 
 & \textbf{\fontsize{10pt}{10pt}\selectfont Helpfulness} & \textbf{\fontsize{10pt}{10pt}\selectfont Personalization} & \\ \hline
                Conversations RAG & 8.07  & 7.48 & - & 2.89 \\ 
                No Persona & 7.96 & 7.35 & - & 2.24 \\ 
                Golden Persona & 8.34 & 7.78 & - & 1.78 \\ \hline
                Persona Learning &  &  &  &  \\ 
                - k=1 & 8.09 & 7.59 & 5.88 & 1.98 \\ 
                - k=3 & 8.29 & 7.63 & 6.07 & 1.81 \\ 
                - k=5 & 8.03 & 7.59 & 5.23 & 2.15 \\ \hline
\end{tabular}
}
\label{tab:main}
\end{table*}

\begin{table*}[ht]
\centering
\caption{Personalized response scores evaluated across base LLMs in three persona settings. The score in \textcolor{red}{red} denotes the improvement of \textit{Persona Learning} over \textit{No Persona} setting.}
\begin{tabular}{l|cccccc}
\hline
\multirow{2}{*}{\textbf{Setting}} & \multicolumn{2}{c}{\textbf{Golden}} & \multicolumn{2}{c}{\textbf{No Persona}} & \multicolumn{2}{c}{\textbf{Persona Learning}} \\
 & \textbf{\fontsize{10pt}{10pt}\selectfont Helpful} & \textbf{\fontsize{10pt}{10pt}\selectfont Personal} 
 & \textbf{\fontsize{10pt}{10pt}\selectfont Helpful} & \textbf{\fontsize{10pt}{10pt}\selectfont Personal} 
 & \textbf{\fontsize{10pt}{10pt}\selectfont Helpful} & \textbf{\fontsize{10pt}{10pt}\selectfont Personal}   \\ \hline
GPT-4o (full bench) & 8.34 & 7.78 & 7.96 & 7.35 & 8.29 {\scriptsize{\color{red}$\triangle$0.33}} & 7.63 {\scriptsize{\color{red}$\triangle$0.28}} \\ \hline
GPT-4o-mini & 8.14 & 7.61 & 8.06 & 7.38 & 8.26 {\scriptsize{\color{red}$\triangle$0.20}} & 7.56 {\scriptsize{\color{red}$\triangle$0.18}} \\ 
Gemini-1.5-pro & 8.16 & 7.93 & 8.17 & 7.37 & 8.27 {\scriptsize{\color{red}$\triangle$0.10}} & 7.64 {\scriptsize{\color{red}$\triangle$0.27}} \\ 
Gemini-1.5-flash & 8.03 & 7.65 & 7.58 & 7.24 & 8.07 {\scriptsize{\color{red}$\triangle$0.49}} & 7.29 {\scriptsize{\color{red}$\triangle$0.05}} \\ 
Claude-1.5-sonnet & 8.11 & 7.28 & 8.01 & 7.11 & 8.03 {\scriptsize{\color{red}$\triangle$0.02}} & 7.20 {\scriptsize{\color{red}$\triangle$0.09}} \\  \hline
\end{tabular}
\label{tab:learn}
\end{table*}



\paragraph{Model Setting}
In our experiments, we conduct persona learning experiments using \textsc{PersonaBench} under three distinct persona settings: \textit{No Persona}, \textit{Golden Persona}, and \textit{Persona Learning}. The experiments were carried out using various proprietary LLMs, specifically the GPT series (gpt-4o, gpt-4o-mini), the Gemini series (gemini-1.5-pro-latest, gemini-1.5-flash-latest), and Claude (claude-3.5-sonnet). To ensure a fair comparison, we use the same prompt template across all models.

To evaluate the effectiveness and robustness of our proposed method, we conducted an ablation study on a randomly selected subset of 10 personas from \textsc{PersonaBench} for all models except gpt-4o, which was evaluated on the entire benchmark.

\subsection{Main Results}

As shown in the top row of Table \ref{tab:main}, the \textit{Golden Persona} setting achieves the highest scores in both personalized response helpfulness (8.34) and personalization (7.78), representing the upper bound of performance for a personalized chatbot. In contrast, the \textit{No Persona} setting serves as a bottom line, where the chatbot makes responses without any prior knowledge of the user's persona. The \textit{Conversations RAG} setting shows a slight improvement in both helpfulness and personalization but with significantly lower utterance efficiency. By examining the results, we observe that the retrieved conversations occasionally cause the personalized chatbot to misinterpret the user's current intent, leading to nuanced responses that may also confuse the user simulator.

Next, we present the results for the AI Persona framework (\textit{Persona Learning} setting) in the bottom row of the table. Specifically, we experiment with different persona update frequencies (\(k=1, 3, 5\). Among these, we observe that updating the persona every 3 conversations yields the best results, with a personalized response helpfulness of 8.29 and a personalization score of 7.63, which are very close to the \textit{Golden Persona} scores. The other two update frequency settings also show slight improvement compared to the \textit{No Persona} setting and \textit{Conversation RAG} setting, demonstrating the effectiveness of our AI Persona framework. 

In terms of the Utterance Efficiency, we can see that in each update frequency, \textit{Persona Learning} shows a remarkable improvement over the \textit{No Persona} baseline. Specifically, the persona update frequency of every 3 conversations (\(k=3\)) results in the best utterance efficiency, closely approaching the performance of the \textit{Golden Persona} setting, indicating the ability of our AI Persona framework to generate relevant and succinct responses that tailored to the user's intent in fewer turns. 

In terms of persona similarity, the \(k=3\) setting exhibits the highest persona similarity score of 6.07, while the \(k=1\) and \(k=5\) settings achieve scores of 5.88 and 5.23, respectively. Notably, this comparison highlights an interesting finding: more frequent updates (as in the \(k=1\) setting) and more information in each update (as in the \(k=5\) setting) do not necessarily result in better learning outcomes.  This finding emphasizes the importance of carefully selecting the learning frequency \(k\) in the life-long personalization of LLMs.

\begin{figure}[ht]
    \centering
    \includegraphics[width=0.8\linewidth]{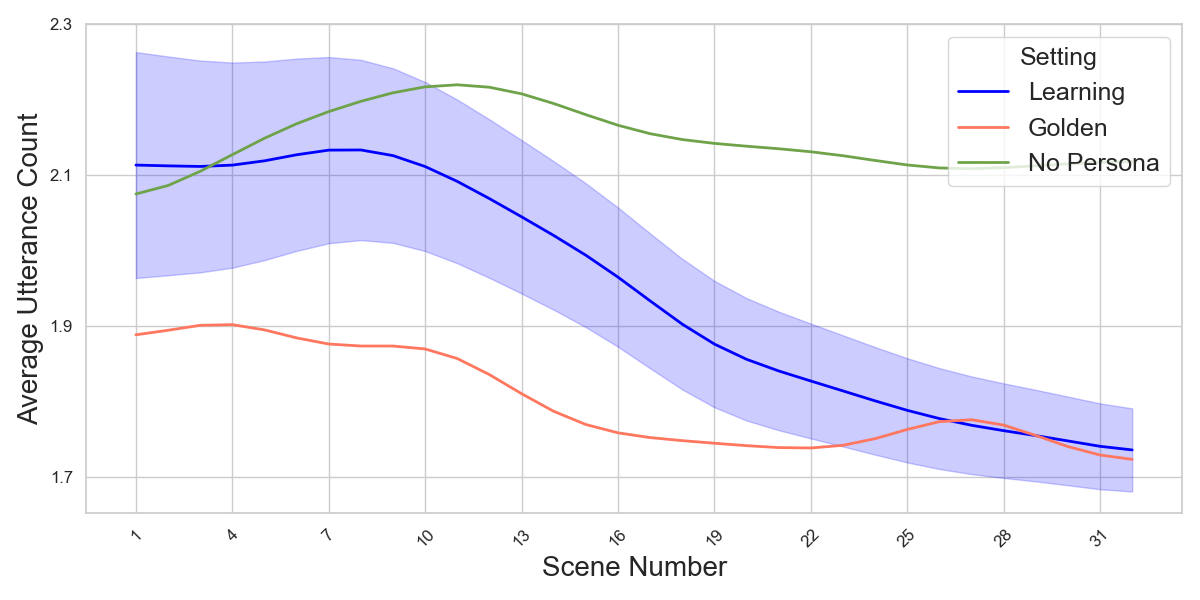}
    \caption{Average number of utterances required per scene. The \textcolor{blue}{blueline} represents \textit{Persona Learning}, the \textcolor[HTML]{ff775f}{orangeline} represents \textit{Golden Persona}, and the \textcolor[HTML]{6ea34a}{greenline} represents \textit{No Persona}. Lower average utterance counts indicate better performance, as it means the dialogue is more efficient and the model requires fewer turns to satisfy the user.}
    \label{fig:Learning Curve}
\end{figure}

\begin{figure}[ht]
    \centering
    \includegraphics[width=0.8\linewidth]{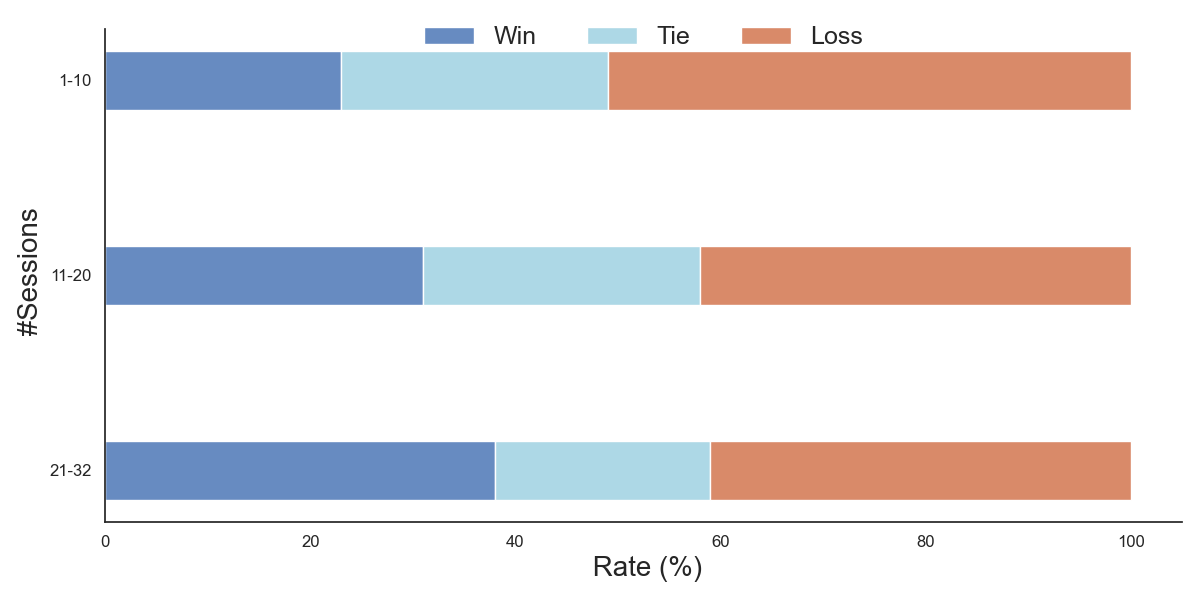}
    \caption{Average winning rate of the pair-wise comparison of \textit{Golden Persona} and \textit{Persona Learning} as the scene number increases.}
    \label{fig:Win Rate}
\end{figure}

\begin{figure}[h]
    \centering
    \includegraphics[width=0.48\textwidth]{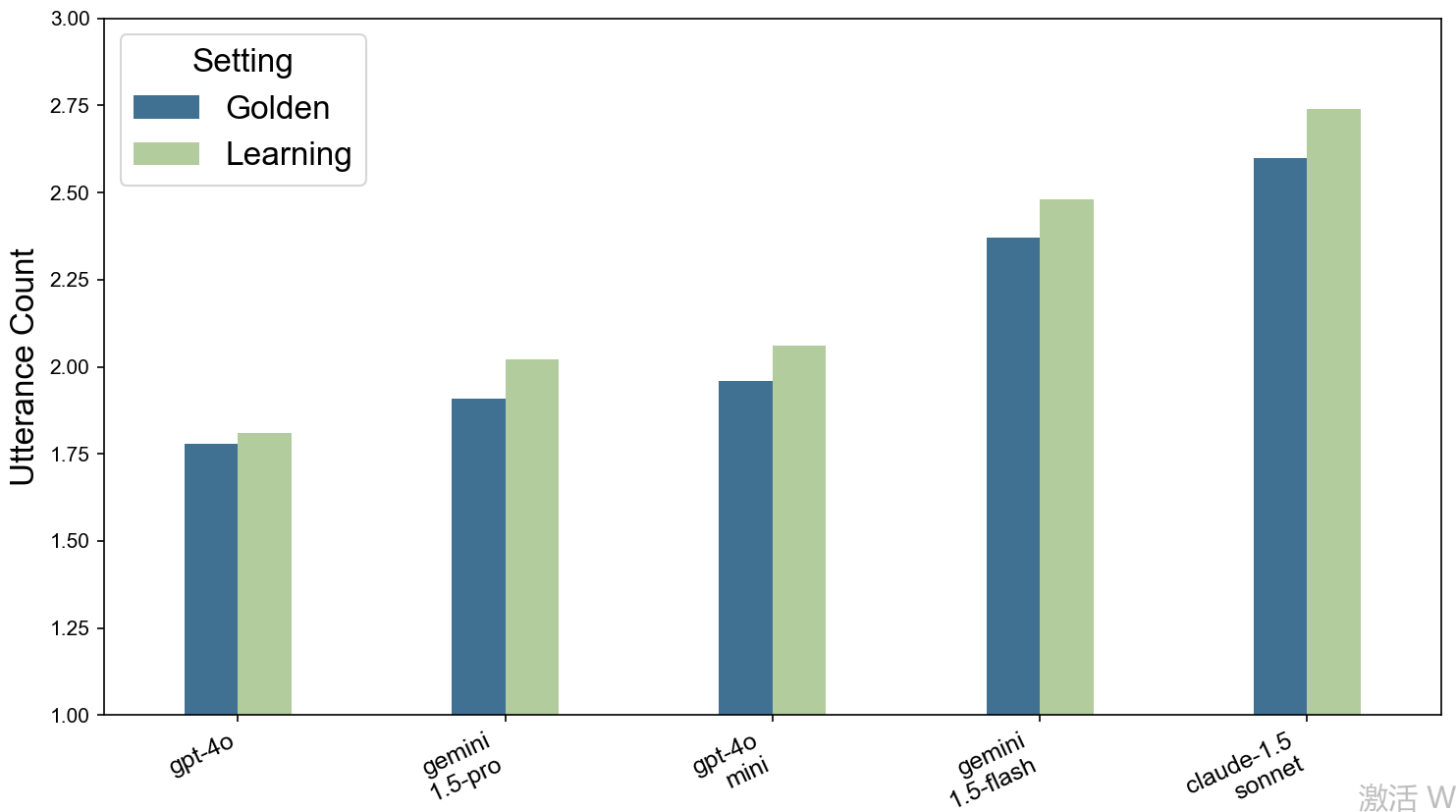} 
    \caption{Average number of utterances for different model bases and persona settings.}  
    \label{fig: Utterances} 
\end{figure}

\subsection{Procedural Learning of Personalized LLMs}
In this subsection, we illustrate how the performance evolves over time as user-agent interactions progress under different settings: \textit{No Persona Access}, \textit{Ground Truth Persona Access}, and \textit{Persona Update}. The primary goal is to evaluate whether the personalized LLM can improve its responses as it engages in more conversations and accumulates more personal information of the user.

As depicted in Figure~\ref{fig:Learning Curve}, the x-axis represents the number of sessions, while the y-axis denotes the number of utterances for the user-simulator to be satisfied. We plot three lines, each corresponding to one of the settings. The comparison demonstrates how the availability of persona information and updating mechanisms influences the model's ability to generate more tailored responses over time.

The \textcolor{blue}{blue learning curve} is the \textit{Persona Learning} setting which showcases a notable improvement over \textcolor[HTML]{6ea34a}{\textit{No Persona}}. The line in the figure reflects a steady decrease in the number of utterances needed for satisfaction and the standard deviation over time as well, demonstrating the effectiveness of persona learning. Remarkably, the performance in the final few sessions under the \textit{Persona Learning} setting approaches that of the \textit{Golden Persona} setting, indicating the effectiveness of our proposed method. It shows that with just over ten updates, the model can learn and adapt to the user's persona efficiently.

To further evaluate the effectiveness of our persona learning framework, we conducted a pairwise comparison of responses generated by the \textit{Persona Learning} setting and the \textit{Golden Persona} setting. Figure~\ref{fig:Win Rate} presents the results, categorized by session groups (1-10, 11-20, 21-32). The win rate of the "Persona Learning" responses steadily increases as the session progresses, demonstrating that our AI persona framework effectively learns and adapts to the user's persona over time.

This procedural learning analysis highlights the importance of dynamic persona modeling in personalized LLM systems, emphasizing the advantages of updating user profiles based on cumulative interactions.

\subsection{Performance across various Base LLMs}
The primary focus of Table~\ref{tab:learn} is to observe the performance of each model under different persona settings. We observe that our proposed \textit{Persona Learning} method improves the performance across all LLMs. Among them gpt-4o shows the best adaptability of personalization.

Figure~\ref{fig: Utterances} summarize the results, demonstrating the average number of utterances and the personalized response scores across different configurations.

The performance differences across base models illustrate that while all LLMs can benefit from persona learning, GPT-4o and Gemini-1.5-pro are better equipped with the ability to adapt and perform in personalized scenarios.

\section{Conclusion}
This paper introduces the task of life-long personalization for large language models (LLMs) and proposes the \textsc{AI Persona} framework, which enables scalable and dynamic adaptation to evolving user's persona without requiring model retraining. We present \textsc{PersonaBench}, a synthesized but realistic and diverse benchmark for evaluating personalized LLMs. Experimental results demonstrate the effectiveness of our framework in improving personalized responses and maintaining updated user profiles. Our work provides a novel, generalizable, and efficient solution for continuous LLM personalization, addressing key limitations in existing approaches.

\section{Limitations}
Although our proposed \textsc{AI Persona} framework are designed to be language-agnostic, the seed data collection and annotation processes in this study were conducted by Chinese native speakers. As a result, the \textsc{PersonaBench} is more representative of scenarios and linguistic nuances specific to Chinese users. While our approach can theoretically generalize to other languages and cultural contexts, its current implementation and evaluation are better suited for Chinese language applications. Future work should involve expanding the data collection and annotation processes to include diverse linguistic and cultural backgrounds to fully validate the framework’s adaptability across different languages and user demographics.
\bibliography{main}
\clearpage
\appendix

\onecolumn 

\vspace{-2cm}
\section{Prompt Templates\label{appb:prompt-template}}
\vspace{-2mm}
\phantomsection
\label{appb:prompt-template9}
\begin{tcolorbox}[width=\textwidth, colback=white!95!gray, colframe=gray!65!black, rounded corners, title={Api call prompt template (Chinese).}]
\begin{CJK*}{UTF8}{gbsn}
\textbf{System Prompt:}

我需要你模拟一个在终端上部署的AI助手，你可以访问、操作终端上的应用，以及进行联网搜索等等。当用户给你输入一个API调用时，我需要你去模拟一下真实的API来返回结果（也就是，假设这个API调用被执行了，会拿到什么样的结果）。  

下面是一些常见的API调用的描述，希望你可以理解这些API的功能，方便你更好地模拟这个API的结果。

\textbf{User Prompt:}  

\{api\_call\}  
\end{CJK*}
\end{tcolorbox}

\vspace{-5mm}
\phantomsection
\label{appb:prompt-template10}
\begin{tcolorbox}[width=\textwidth, colback=white!95!gray, colframe=gray!65!black, rounded corners, title={Api call prompt template (English).}]
\textbf{System Prompt:}

I need you to simulate an AI assistant deployed on a terminal. You can access and operate terminal applications, as well as perform online searches. When the user provides you with an API call, I need you to simulate the actual API and return a result (i.e., assume the API call was executed and provide the kind of result it would return).  

Below are descriptions of some common API calls. Please understand these API functionalities to better simulate the results of these APIs.   

\textbf{User Prompt:}  

\{api\_call\}  

\end{tcolorbox}

\vspace{-1.5mm}
\phantomsection
\label{appb:prompt-template1}
\begin{tcolorbox}[width=\textwidth, colback=white!95!gray,colframe=gray!65!black,rounded corners, title={Personalized chatbot API call prompt template(Chinese).}]
\begin{CJK*}{UTF8}{gbsn}
\textbf{System Instruction: }

你是一个在终端上部署的AI助手，除了像ChatGPT一样的多轮对话之外，你还可以访问、操作终端上的应用，以及进行联网搜索。

\textbf{场景描述：}  

\{scene\}  

在这个场景中你可能需要访问或调用的API有：  

\{api\_docs\}  

\textbf{用户的人设信息和使用习惯如下：}  

\{persona\}  

根据你对当前场景和用户提问的理解，你可以自行决定是否需要访问或调用API。  

如果你认为要给出一个好的回复，你必须要调用API，则请你显式地在回复中使用形如下面的api调用的例子来进行回复：  

\{API\_Example\}  

并且将这个回复放在最开头。如果有部分内容是可以不访问/调用API就可以回答的，你可以先尽可能回答问题，再在最后补充类似于关于\{具体需求\}，我需要访问/调用\{具体API名称\}才能提供完整的答案。然后再参照api call的例子中的格式输出你的API call即可。

\textbf{User Prompt:}

\{query\}
\end{CJK*}
\end{tcolorbox}

\phantomsection
\label{appb:prompt-template2}
\begin{tcolorbox}[width=\textwidth, colback=white!95!gray,colframe=gray!65!black,rounded corners, title={Personalized chatbot API call prompt template (English).}]
\textbf{System Instruction: }

You are an AI assistant deployed on a terminal. In addition to multi-turn conversations like ChatGPT, you can also access and operate applications on the terminal and perform online searches.

Scenario Description:

\{scene\}  

In this scenario, you may need to access or call the following APIs:  

\{api\_docs\}  

User persona information and usage habits: 

\{persona\}  

Based on your understanding of the current scenario and the user's query, you can decide whether you need to access or call an API.  

If you believe that providing a good response requires calling an API, you must explicitly include the API call in your response using the example format below:  

\{API\_Example\}  

Make sure this response is placed at the very beginning. If part of the query can be answered without accessing/calling an API, try to answer that first. Then, at the end, add a statement such as regarding specific need, I need to access/call specific API name to provide a complete answer. Finally, include the API call in the example format as a supplement.

\textbf{User Prompt:}

\{query\}
\end{tcolorbox}

\phantomsection
\label{appb:prompt-template3}
\begin{tcolorbox}[width=\textwidth, colback=white!95!gray,colframe=gray!65!black,rounded corners, title={Persona Update prompt template(Chinese).}]
\begin{CJK*}{UTF8}{gbsn}
\textbf{System Instruction: }

你是一个用于用户画像抽取的AI机器人，你的任务是基于用户的人设config和用户最近使用AI助手的对话历史来判断人设的哪些字段（field）需要被更新。

\textbf{当前的用户画像（人设config）如下:}  

\{persona\}  

注意，我需要你仔细斟酌在上面的对话历史中，思考用户的人设有没有哪里发生了改变。  你可以先理解人设、分析近期的对话，再来判断有哪些字段是需要被更新的。  

最终的待更新的字段用如下的形式输出：  

\textless fields\textgreater

这里是具体要更新的字段和字段更新后的内容 

\textless/fields\textgreater 

下面是一个输出的例子:  

\{Fields\_Update\_Example\}  

\textbf{User Prompt:} 

\{chat\_history\}
\end{CJK*}
\end{tcolorbox}

\phantomsection
\label{appb:prompt-template4}
\begin{tcolorbox}[width=\textwidth, colback=white!95!gray, colframe=gray!65!black, rounded corners, title={Persona Update prompt template (English).}]
\textbf{System Instruction: }

You are an AI bot designed for extracting user personas. Your task is to determine, based on the user's persona config and their recent conversation history with the AI assistant, which fields (if any) in the persona need to be updated.

Current user persona config:

\{persona\}  

Note: Please carefully consider the conversation history above and reflect on whether any changes have occurred in the user's persona. You can start by understanding the persona and analyzing recent conversations before determining which fields need to be updated.

You can first understand the persona and analyze the recent conversation before determining which fields need to be updated.  

The fields to be updated should be output in the following format:  

\textless fields\textgreater

Here are the specific fields that need updating and their updated content.  

\textless/fields\textgreater

Below is an example of the output:  

\{Fields\_Update\_Example\}  

\textbf{User Prompt:}

\{chat\_history\}
\end{tcolorbox}

\phantomsection
\label{appb:prompt-template5}
\begin{tcolorbox}[width=\textwidth, colback=white!95!gray, colframe=gray!65!black, rounded corners, title={User Simulation prompt template (Chinese).}]
\begin{CJK*}{UTF8}{gbsn}
\textbf{System Prompt:}

接下来我需要你扮演一个真实的人类来和此人常用的AI助手进行多轮对话。我会给你一个persona setting，请你先理解这个persona的内容再将自己完全代入到这个角色当中。

\textbf{给你的用户画像如下：}  

姓名：\{name\}  

年龄：\{age\}  

性别：\{gender\}  

国籍：\{nationality\}  

语言：\{language\}  

职业信息：\{career\}  

MBTI: \{MBTI\}  

价值观与爱好：\{values\}  

行为画像：\{pattern\}  

使用偏好：\{preference\}

\textbf{当前场景描述：} 

\{scene\}。

还有些更有语境的场景信息可以供你参考，（注意！只是参考，不要全部使用！）：  

\{scene\_context\}  

下面是参考的例子: 

\{EXAMPLE\}  

注意事项：
1. 这里的AI不是一个概念型AI（钢铁侠中的Jarvis那种什么都能做的就是概念型AI），而是一个终端上的AI助手，我们假设AI的功能就只有终端上的应用级操作、联网搜索和正常的多轮对话。  
2. 请你牢记你的AI助手是非常了解你的，所以你无需在提问时重申你的人设或背景情况，当且仅当你发现AI助手给你的回复并不符合你的人设或场景设定时，你才能在后续的聊天中补充你的需求。  
3. 在模拟对话的时候我需要你能够真实带入到AI助手的使用者的这个视角，语气不要太客气，同时不要主动的询问你的AI助手“你还需要哪些信息”等等，要让AI助手来询问你，你再给出相应的信息。  

一个可以供你参考的流程为：  
先简单描述一下问题，并说出需要AI助手帮你做的事，然后等待AI给出正确的理解（如果发现AI没有正确理解你的意图，则需要再次纠正）之后，再配合AI给出相关信息。  

我相信你已经get到了，现在，无须说任何多余的废话，立即进入你的角色！

\textbf{User Prompt:}  

\{chat\_history\}
\end{CJK*}
\end{tcolorbox}

\phantomsection
\label{appb:prompt-template6}
\begin{tcolorbox}[width=\textwidth, colback=white!95!gray, colframe=gray!65!black, rounded corners, title={User Simulation prompt template (English).}]
\textbf{System Prompt:}

You will now play the role of a real human engaging in multi-turn conversations with their commonly used AI assistant. I will provide you with a persona setting. Please first understand the persona details and fully immerse yourself into this role.

User Persona Setting:

Name: \{name\}  

Age: \{age\}  

Gender: \{gender\}  

Nationality: \{nationality\}  

Language: \{language\}  

Career Info: \{career\}  

MBTI: \{MBTI\}  

Values and Hobbies: \{values\}  

Behavioral Traits: \{pattern\} 

Usage Preferences: \{preference\}  

Current Scenario Description:

\{scene\}. 

Additional contextual scene information is provided for reference (Note: Use it as reference only, do not fully adopt it!): 

\{scene\_context\}  

The following is an example for reference: 

\{EXAMPLE\}  

\textbf{Important Notes:}  
1. This AI is not a conceptual AI (like Jarvis in Iron Man, which can handle everything), but rather a terminal-based assistant. Assume its functionalities are limited to terminal operations, online searches, and normal multi-turn dialogues.  
2. Keep in mind that the AI assistant knows you very well. Thus, you don't need to reiterate your persona or background when asking questions. Only if the AI's response does not align with your persona or scenario should you provide additional clarifications.  
3. When simulating the dialogue, fully immerse yourself in the perspective of the AI assistant user. Avoid being overly polite, and do not proactively ask the AI questions like, "What other information do you need?" Let the AI prompt you for additional details instead.  

 A typical flow to follow: 
First, briefly describe your issue and what you need the AI to do. Wait for the AI's understanding (and correct its interpretation if needed). Then provide the necessary details to proceed.  

I believe you've got it. Now, without saying anything unnecessary, immediately step into your role!

\textbf{User Prompt:}  

\{chat\_history\}

\end{tcolorbox}

\begin{CJK*}{UTF8}{gbsn}

\phantomsection
\label{appb:prompt-template7}
\begin{tcolorbox}[width=\textwidth, colback=white!95!gray, colframe=gray!65!black, rounded corners, title={Satisfaction check prompt template (Chinese).}]
\textbf{System Prompt:}

接下来我需要你扮演一个真实的人类，我会给你一个人设的persona，请你先理解这个persona的内容再将自己完全代入到这个角色当中。

\textbf{给你的用户画像如下：}  

姓名：\{name\}  

年龄：\{age\}  

性别：\{gender\}  

国籍：\{nationality\}  

语言：\{language\}  

职业信息：\{career\}  

MBTI: \{MBTI\}  

价值观与爱好：\{values\}  

行为画像：\{pattern\}  

使用偏好：\{preference\}

请你先理解上面给出的人设和场景，再根据你的理解去检查以下对话是否达到了“你”（即这个人设）预期的目标。  

\textbf{对话记录：}  

\{chat\_history\}  

下面是另一个 Personalized Agent 对当前人设和场景给出的预期应答，你可以将其作为参考，但是最终对于是否符合要求还是由你来定夺。  

\{expected\_results\}  

\textbf{注意：}  

你的输出必须遵循以下原则：  
如果认为 AI 助手回答的可以了，就输出且只输出：<满意>；  
如果你认为回答不符合要求的话，请你输出且只输出：<继续>。  

\textbf{User Prompt:}  

\{chat\_history\}

\end{tcolorbox}
\end{CJK*}

\phantomsection
\label{appb:prompt-template8}
\begin{tcolorbox}[width=\textwidth, colback=white!95!gray, colframe=gray!65!black, rounded corners, title={User Simulation prompt template (English).}]
\textbf{System Prompt:}

Next, I need you to play the role of a real human. I will provide you with a persona setting. Please first understand the persona's content and fully immerse yourself in this role.

\textbf{User Persona Setting:}  

Name: \{name\}  

Age: \{age\}  

Gender: \{gender\}  

Nationality: \{nationality\}  

Language: \{language\}  

Career Info: \{career\}  

MBTI: \{MBTI\}  

Values and Hobbies: \{values\}  

Behavioral Traits: \{pattern\} 

Usage Preferences: \{preference\}  

Please first understand the above persona and scenario. Then, based on your understanding, evaluate whether the following conversation meets the expectations of "you" (i.e., this persona).  

Conversation History:

\{chat\_history\}  

Below is the expected response given by another Personalized Agent for the current persona and scenario. You can use it as a reference, but ultimately, it is up to you to decide whether it meets the requirements.  

\{expected\_results\}  

\textbf{Note:}  

Your output must follow the principle below:  
If you think the AI assistant's response is acceptable, output and only output: <Satisfied>.  
If you think the response does not meet the requirements, output and only output: <Continue>.  

\textbf{User Prompt:}  

\{chat\_history\}

\end{tcolorbox}












\phantomsection
\label{appb:prompt-template11}
\begin{tcolorbox}[width=\textwidth, colback=white!95!gray, colframe=gray!65!black, rounded corners, title={Prompt Template (GPT Evaluation for Personalized Responses, Chinese).}]
\begin{CJK*}{UTF8}{gbsn}
\textbf{System Prompt:}

接下来我会给你一个用户和AI助手的对话以及这个用户的人设Persona信息。我需要你先理解用户的Persona，再基于用户这一轮的Query和AI助手给出的回答来打分，以评估AI助手的回答在多大程度上解决了用户问题中表达的顾虑和需求。  

\textbf{评分维度如下：}  

1. \textbf{问题解决程度}  

10分：AI助手完全满足了用户需求，并提供了清晰、有效且有帮助的解决方案。  

7分：AI助手部分满足了用户需求，但可能存在一些不足，例如缺少方案的具体信息、API接口的调用或流程的操作步骤。  

5分：AI助手提供了一些与用户需求相关的知识，但没有真正解决问题。 

3分：AI助手提供的知识与用户需求没有直接关系，或信息缺乏准确性。 

0分：AI助手没有提供任何与用户需求相关的解决方案或信息。  

2. \textbf{个性化程度}  

10分：AI助手完全理解用户的Persona，并根据用户的特点和喜好，提供了完整的个性化的回复和建议。  

7分：AI助手在回复中体现了对用户Persona的部分理解，但仍有进一步提升的空间。  

5分：AI助手没有明显体现对用户个性的理解，只是单纯地进行了对话型问答。  

0分：AI助手对用户Persona的理解错误，误解了用户的意图，进行了错误的回答。  

\textbf{输出格式：}  

\textless analysis\textgreater  

这里针对 solution\_score 和 persona\_score 可以做出分析。  

\textless/analysis\textgreater  

\textless rating\textgreater  

solution\_score; persona\_score  

\textless/rating\textgreater  

\textbf{User Prompt:}  

以下是用户画像，用户的Query和AI助手的回答： 

用户画像：\{persona\}  

用户Query：\{query\} 

AI助手回答：\{answer\}  
\end{CJK*}
\end{tcolorbox}

\phantomsection
\label{appb:prompt-template12}
\begin{tcolorbox}[width=\textwidth, colback=white!95!gray, colframe=gray!65!black, rounded corners, title={Prompt Template (GPT Evaluation for Personalized Responses, English).}]
\textbf{System Prompt:}

I will now provide you with a conversation between a user and an AI assistant, along with the persona information of the user. You are required to first understand the user's persona and then evaluate the AI assistant's response based on the user's query and the assistant's response.  

Evaluation Dimensions:

1. Solution Score

10: The AI assistant fully meets the user's needs and provides clear, effective, and helpful solutions.  

7: The AI assistant partially meets the user's needs but has certain shortcomings, such as missing detailed information, API calls, or operational steps.  

5: The AI assistant provides some knowledge related to the user's needs but does not solve the problem.  

3: The AI assistant provides knowledge unrelated to the user's needs or lacks accuracy.  

0: The AI assistant provides no relevant solutions or information related to the user's needs.  

2. Personalization Score

10: The AI assistant fully understands the user's persona and provides completely personalized responses and suggestions based on the user's characteristics and preferences.  

7: The AI assistant shows partial understanding of the user's persona in its response but has room for improvement.  

5: The AI assistant shows no clear understanding of the user's persona and merely performs generic Q\&A.  

0: The AI assistant misunderstands the user's persona and provides incorrect responses that do not align with the user's intent.  

Output Format:

\textless analysis\textgreater  

An analysis can be conducted on solution\_score and persona\_score.

\textless/analysis\textgreater  

\textless rating\textgreater  

solution\_score; persona\_score  

\textless/rating\textgreater  

\textbf{User Prompt:}  

Below are the user's persona, query, and the AI assistant's response:  

User Persona: \{persona\}  

User Query: \{query\}  

AI Assistant Response: \{answer\}  

\end{tcolorbox}

\phantomsection
\label{appb:prompt-template13}
\begin{tcolorbox}[width=\textwidth, colback=white!95!gray, colframe=gray!65!black, rounded corners, title={Prompt Template (GPT Evaluation for Persona Similarity, Chinese).}]
\begin{CJK*}{UTF8}{gbsn}
\textbf{System Prompt:}

接下来我会给你一个用户的两个Persona人设信息，一个是\textless ground\_truth\textgreater 人设，另一个是\textless learned\_persona\textgreater 人设。  
\textless learned\_persona\textgreater 是通过一套外部的Persona Learning Pipeline基于用户和AI助手在不同场景中的对话来建模的特征。  
我需要你先阅读\textless ground\_truth\textgreater 里的Persona，并理解这个用户的真实人设。再去给\textless learned\_persona\textgreater 的人设整体的一致性和人设的细节还原程度打分。  

\textbf{评分维度如下：}  

1. \textbf{人设整体的一致性打分：}  

用来评估生成的人设是否在整体语义上与\textless ground\_truth\textgreater 保持一致，包括语义的相关性性和不同字段内容的连贯性。  

10分：整体语义完全一致，行为连贯，无歧义或偏差。  

7分：整体语义基本一致，但存在少量不影响理解的小差异。  

5分：整体语义部分一致，但有明显差异，影响整体理解。  

3分：整体语义偏离较大，信息传达困难或有较大矛盾。  

0分：完全不一致，语义严重背离目标人设。  

2. \textbf{细节还原度打分：}  

用来评估模型对于各个字段建模出来的是否准确、细致地还原了目标人设中的各个字段，包括关键特征和相关细节。  

10分：所有字段的内容高度准确、具体，完整覆盖目标人设的特征和细节。  

7分：大部分字段还原准确，细节基本到位，但有少量字段略显模糊或缺失。 

5分：部分字段能体现用户特征，但整体还原不够深入，细节缺失较多。 

3分：多数字段还原度较低，内容笼统，或字段描述存在错误。  

0分：字段内容与目标人设完全不符，大量细节缺失或错误。  

\textbf{输出格式：}  

\textless analysis\textgreater  

这里针对\textless learned\_persona\textgreater 和\textless ground\_truth\textgreater 整体内容和个别字段内容可以做出分析。  

\textless/analysis\textgreater  

\textless rating\textgreater  

consistency\_score ; detail\_restoration\_score  

\textless/rating\textgreater  

\textbf{User Prompt:}  

下面是真实用户人设：  

\textless ground\_truth\textgreater  

\{persona\_gt\}  

\textless/ground\_truth\textgreater  

下面是通过人机交互的对话历史建模到的人设：  

\textless learned\_truth\textgreater 

\{persona\_learned\}  

\textless/learned\_truth\textgreater  
\end{CJK*}
\end{tcolorbox}

\phantomsection
\label{appb:prompt-template14}
\begin{tcolorbox}[width=\textwidth, colback=white!95!gray, colframe=gray!65!black, rounded corners, title={Prompt Template (GPT Evaluation for Persona Similarity, English).}]
\textbf{System Prompt:}

I will provide you with two persona profiles for a user: one is the \textless ground\_truth\textgreater persona, and the other is the \textless learned\_persona\textgreater persona.  
The \textless learned\_persona\textgreater is modeled based on user-AI interactions in different scenarios through an external Persona Learning Pipeline.  
You need to first read the persona in \textless ground\_truth\textgreater and understand the user's true characteristics. Then, score the \textless learned\_persona\textgreater\ for overall consistency and detail restoration.  

Scoring Dimensions:

1. Overall Consistency Score:

Used to evaluate whether the generated persona is semantically consistent with the \textless ground\_truth\textgreater\ persona, including semantic relevance and coherence across different fields.  

10: Fully consistent in semantics, coherent behavior, no ambiguity or deviation.  

7: Mostly consistent in semantics, with minor differences that do not affect understanding.  

5: Partially consistent, with noticeable differences that affect overall understanding.  

3: Significant deviation in semantics, difficult to understand or conflicting information.  

0: Completely inconsistent, with severe semantic deviation from the target persona.  

2. Detail Restoration Score: 

Used to evaluate whether the fields in the model accurately and thoroughly restore the target persona, including key characteristics and relevant details.  

10: All fields are highly accurate, specific, and fully cover the target persona's features and details.  

7: Most fields are accurately restored with basic details in place, but some fields are slightly vague or missing.  

5: Some fields reflect user characteristics, but overall restoration lacks depth, with many missing details.  

3: Most fields have low restoration accuracy, with vague content or incorrect descriptions.  

0: Field content does not match the target persona, with substantial missing or incorrect details.  

Output Format:

\textless analysis\textgreater  

Provide analysis on the overall content and specific fields of \textless learned\_persona\textgreater compared to \textless ground\_truth\textgreater.  

\textless/analysis\textgreater  

\textless rating\textgreater  

consistency\_score ; detail\_restoration\_score  

\textless/rating\textgreater  

\textbf{User Prompt:}  

Below is the ground-truth persona:  

\textless ground\_truth\textgreater  

\{persona\_gt\}  

\textless/ground\_truth\textgreater  

Below is the persona modeled through user-AI interaction:  

\textless learned\_truth\textgreater  

\{persona\_learned\}  

\textless/learned\_truth\textgreater  

\end{tcolorbox}

\end{document}